\documentclass{article}
\usepackage{ijcai20}

\usepackage{times}
\usepackage{soul}
\usepackage{url}
\usepackage[hidelinks]{hyperref}
\usepackage[utf8]{inputenc}
\usepackage[small]{caption}
\usepackage{graphicx,import}
\usepackage{amsmath}
\usepackage{amsthm}
\usepackage{booktabs}
\usepackage{algorithm}
\usepackage{algorithmic}
\usepackage{shortcuts, wrapfig}
\usepackage{amssymb, parskip,color, setspace,tabularx, wrapfig,float,enumerate} 

\title{Graph Deconvolutional Generation}

\author{
Daniel Flam-Shepherd$^{1,2,*}$
\and
Tony Wu$^{3,*}$\and
Alan Aspuru-Guzik$^{1,2,3,4}$
\affiliations
$^1$Department of Computer Science, University of Toronto, Canada.\\
$^2$Vector Institute for Artificial Intelligence, Toronto, Canada \\
$^3$Department of Chemistry, University of Toronto, Canada \\
$^4$Canadian Institute for Advanced Research (CIFAR) Senior Fellow, Toronto, Canada
\emails
}

\begin{document}

\maketitle

\begin{abstract}
    Graph generation is an extremely important task, as graphs are found throughout different areas of science and engineering. 
    In this work, we focus on the modern equivalent of the Erdos-R\'{e}nyi random graph model: the graph variational autoencoder (GVAE) \cite{simonovsky2018graphvae}. 
    This model assumes edges and nodes are independent in order to generate entire graphs at a time using a multi-layer perceptron decoder. 
    As a result of these assumptions, GVAE has difficulty matching the training distribution and relies on an expensive graph matching procedure.
    We improve this class of models by building a message passing neural network into GVAE's encoder and decoder. 
    We demonstrate our model on the specific task of generating small organic molecules.
\end{abstract}

\section{Introduction}

In the past five years there has been rapid progress in the development of deep generative models for 
continuous data like images \cite{kingma2013auto} and sequences like natural language \cite{sutskever2011generating}).
The two most prominent models developed are generative adversarial networks (GANs) 
\cite{goodfellow2014generative} and variational autoencoders (VAEs) \cite{kingma2013auto}, 
both learn a distribution parameterized by neural networks. 

 There has also been tremendous progress in deep generative models for combinatorial structures, particularly graphs. 
 Graph generation is an important research area with significant applications in drug and material designs. 
 However, discovering new compounds with specific properties is an extremely challenging task 
 because of the huge, unstructured and discrete nature of the search space. 
 Hence, more effort needs to be directed towards building simple yet efficient models with the proper inductive biases. 
 We focus our efforts on this domain specific area of generating molecular graphs. 

 One of the first graph generative models is the Erdos-R\'{e}nyi (ER) random graph model \cite{erdHos1960evolution} where each edge and node exists with independent probability.  
 The modern approach from deep generative models in this class is GVAE where the decoder outputs independent probabilities for edge and node features.

Another class of generative models of graphs are sequential models that construct graphs sequentially, node by node. 
When generating molecules, these models can be constrained to ensure they only generate valid molecules, but they must be trained one molecule at a time since the generative process depends on a sequence of probabilistic decisions. This also makes them significantly more difficult to train and tune. 

Models in the ER family such as GVAE, do not have this requirement and are significantly easier to train.
However, because they do not consider edge correlations, it is much harder to match the training distribution with these models. 
This also makes them difficult if not impossible to constrain so that they only generate valid molecules. 
For example, \cite{ma2018constrained} proposes a regularization framework for GVAEs to in order for them to generate semantically valid graphs.
They formulate penalty terms that address validity constraints. 
However, they only obtain limited success on small graphs and are significantly less successful on larger molecules.

\begin{figure}[t]
\begin{center}
\def\svgwidth{0.9\linewidth}
\begingroup%
  \makeatletter%
  \providecommand\color[2][]{%
    \errmessage{(Inkscape) Color is used for the text in Inkscape, but the package 'color.sty' is not loaded}%
    \renewcommand\color[2][]{}%
  }%
  \providecommand\transparent[1]{%
    \errmessage{(Inkscape) Transparency is used (non-zero) for the text in Inkscape, but the package 'transparent.sty' is not loaded}%
    \renewcommand\transparent[1]{}%
  }%
  \providecommand\rotatebox[2]{#2}%
  \newcommand*\fsize{\dimexpr\f@size pt\relax}%
  \newcommand*\lineheight[1]{\fontsize{\fsize}{#1\fsize}\selectfont}%
  \ifx\svgwidth\undefined%
    \setlength{\unitlength}{573.29783042bp}%
    \ifx\svgscale\undefined%
      \relax%
    \else%
      \setlength{\unitlength}{\unitlength * \real{\svgscale}}%
    \fi%
  \else%
    \setlength{\unitlength}{\svgwidth}%
  \fi%
  \global\let\svgwidth\undefined%
  \global\let\svgscale\undefined%
  \makeatother%
  \begin{picture}(1,0.31405488)%
    \lineheight{1}%
    \setlength\tabcolsep{0pt}%
    \put(0,0){\includegraphics[width=\unitlength,page=1]{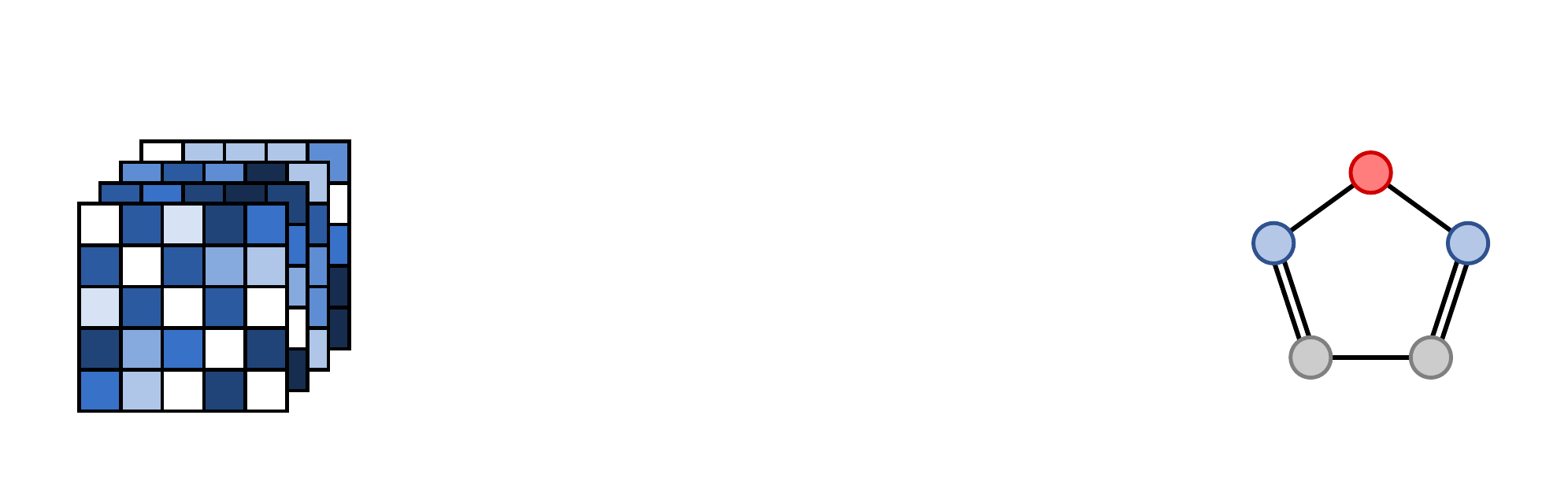}}%
    \put(0.13579815,0.25798405){\color[rgb]{0,0,0}\makebox(0,0)[t]{\lineheight{1.25}\smash{\begin{tabular}[t]{c}\scalebox{1.0}{Edge Features}\end{tabular}}}}%
    \put(0.51007692,0.25798405){\color[rgb]{0,0,0}\makebox(0,0)[t]{\lineheight{1.25}\smash{\begin{tabular}[t]{c}\scalebox{1.0}{Node Features}\end{tabular}}}}%
    \put(0.87272222,0.25798405){\color[rgb]{0,0,0}\makebox(0,0)[t]{\lineheight{1.25}\smash{\begin{tabular}[t]{c}\scalebox{1.0}{Molecular Graph}\end{tabular}}}}%
    \put(0,0){\includegraphics[width=\unitlength,page=2]{figure1_1.pdf}}%
  \end{picture}%
\endgroup%

\end{center}
\caption{The Erdos-Renyi model for c1cnon1 }
\end{figure}

We propose that more effort should be directed towards improving the underlying model design of GVAE before attempting to construct regularization frameworks. 
In this work, we focus on improving GVAE's basic structure: in particular we note that its MLP decoder does not have the proper inductive bias for graph structured data. 
Therefore, we introduce a simple model for graph generation by building a message passing neural network into the encoder and decoder of a VAE (MPGVAE).
Our model is simple to implement but is more adept at domain specific graph generation and even allows us to avoid using graph matching.
We demonstrate the MPGVAE on a few standard tasks in molecular generative modeling.

\newpage

\section{Background}

In this paper we focus on undirected graphs $G$ with $n$ nodes denoted $G= (\B A , \B E, \B X) $, where the adjacency matrix is $\B A \in \{0,1\}^n$ such that $\B A_{uv}=1$ implies nodes $u$ and $v$ are connected.
The edge feature tensor $ \B E \in \R ^{n\times n \times e} $ specifies edge features, with $\B E_{uvw}=1$ denoting that nodes $u$ and $v$ have edge type $w$ with specific edge features $\B e_{uv}\in \R ^e$ . 
The node feature matrix $\ \ \B X \in \R ^ {n \times f }$ stacks all node feature vectors $\B x _v \in \R ^f$.  

\subsection{Neural message passing} 
Message passing neural networks (MPNNs) \cite{gilmer2017neural} operate on graphs $G$. First, a message passing phase runs for $T$ propagation steps and is defined in terms of message functions $M_t$ and node update functions $U_t$.  
During the message passing phase, hidden states $\B h^t_v$ of each node in the graph are updated based on messages $\B m^{t+1}_v$ according to
\begin{align}
 \B m _v ^{t+1} &= \sum _{w\in \N _v } M_t(\B h_v ^t, \B h_w^t, \B e_{vw}) \\
 \B h^{t+1}_v &= U_t(\B h^{t}_v, \B m^{t+1}_v)
\end{align}
Every node $v$ receives an aggregate message from its neighbours $\N_v$, in this case, by simple summation. 
Then in the second phase we readout predictions $\B y$ based on final node embeddings, after $T$ propagation steps. 
\begin{align}
     \B y = \texttt{Readout}(\{\B h^T _v\}_{v\in G})    
\end{align}

\subsection{Variational Auto-Encoders for Graphs }

A GVAE learns a probability distribution from a set of training graphs $\D _G $  such that we can sample new graphs from it. 
To do this, a VAE learns a latent representation $\B z$ of those graphs so that the generative model $p_{\theta }(G|\B z )$, 
which is defined by a neural network with parameters $\theta$, can generate a graph $G$. 
Assuming the training data are independent, the objective is to maximize the log-evidence of the data 
\begin{align}
    \E _{p(\D_G)} [\log p(G)] =\E _{p(\D_{\mathcal{G}})} [\log \E _{\B z \sim p(\B z)} [p(G|\B z)] ],
\end{align}
which is intractable but can be lower-bounded by introducing a variational approximation $q_{\B \phi} (\B z | G)$. 
Another neural network called the inference network encodes a training graph $G$ into a latent representation $\B z$ and outputs the parameters of $q_{\B \phi} (\B z | G)$
\begin{align}
     q(\B z  | G ) = \N (\B z | \mu (G), \sigma (G)).
\end{align}
The latent space $z$ is low dimensional to make sure the model does not just memorize the training data.
We maximize the lower bound with respect to the generator and inference network parameters :
\begin{align}
\log p(G)  \geq   \E _{q_{\B \phi} (\B z | G) }[p_{\theta }(G|\B z )] - \kl [ q_{\B \phi} (\B z | G) || p(\B z) ]    
\end{align}
where $\kl $ denotes the Kullback–Leibler divergence. 
The first term in (6) is the reconstruction loss which ensures  
the generated graphs are similar to the training graphs.
The second term in (6) regularizes the latent space
to ensure that we spread the density of the generative model around.
The prior $p(\B z)$ is a standard normal.
\newpage 

\section{Related Work}

\textbf{Graph neural networks.} The first neural network operating on graphs model was proposed by \cite{scarselli2008graph}, 
and later improved upon by \cite{li2015gated}.
A general framework based on neural message passing was constructed by \cite{gilmer2017neural}. 
Also, \cite{DuvMacetal15nfp} introduced a convolutional neural network for molecular graphs.

\textbf{Generating SMILES.} Several works have explored training generative models on SMILES representations of molecules. One of the first, CharacterVAE \cite{gomez2018automatic} is based on a VAE with recurrent neural networks. The GrammarVAE \cite{kusner2017grammar} and SDVAE
\cite{dai2018syntax} constrain the decoder in order to follow particular syntactic and semantic rules.
Recently, these works were extended by \cite{krenn2019selfies} to ensure 100\% reconstruction validity. 

\textbf{Sequential models.} The first of such models comes from 
\cite{johnson2016learning}, which incrementally constructs a graph as a sequence of probabilistic decisions, in order to do some reasoning task. Using this framework, DeepGMG \cite{li2018learning} built an autoregressive model for graphs conditioned on the full generation history. CGVAE \cite{liu2018constrained} improves upon DeepGMG by building a Gated Graph NN \cite{li2015gated} into the encoder and decoder of a VAE \cite{kingma2013auto}, but it only conditions on the current partial graph during generation. Graph convolution policy network \cite{you2018graph} uses a GCN \cite{kipf2016semi} to sequentially generate molecules in goal-directed way through reinforcement learning. In general, sequential models have some issues with stability and scaling as larger graphs require more propagation steps.

More generally GraphRNN \cite{you2018graphrnn} generates the adjacency matrix sequentially, 
one entry or one column at a time through an RNN. However, the GraphRNN scales quadratically in number of nodes and can't handle long term dependencies. A recent work, improves upon this, GRAN \cite{liao2019efficient} generates graphs one block of nodes and associated edges at a time.

\textbf{Modern ER models.} These are models which generate entire graphs at a time where each edge and node is independent; these include Graphvae \cite{simonovsky2018graphvae} and MolGAN \cite{de2018molgan}   
which avoids issues arising from node ordering that are associated with likelihood based methods by using an adversarial loss instead
\cite{goodfellow2014generative}. 

\textbf{MPGVAE as domain specific Graphite.} Graphite \cite{grover2018graphite} 
is a recently introduced method for generative model of graphs parameterizing a
variational auto-encoder with a graph neural network using a iterative graph 
refinement strategy in the decoder. This particular work is very similar to MPGVAE but 
uses a simple GCN \cite{kipf2016semi} and includes latent variables per node, 
our model can be seen as a domain specific version of Graphite for more complicated constrained graphs.

\begin{figure*}
\begin{center}
\def\svgwidth{0.85\linewidth}
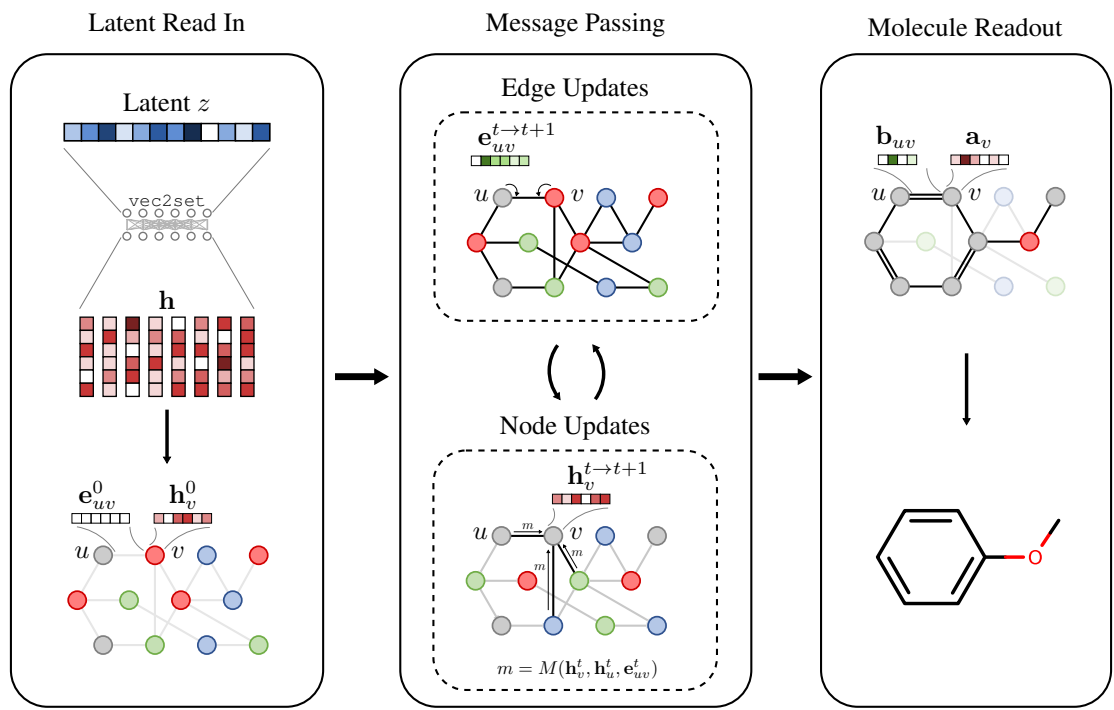
\end{center}
\caption{The Generative Process for MPGVAE 1) first we read in an unconnected graph from the fixed dimension latent space then 2) we perform message passing on the raw graph creating both an edge and node representation to 3) read out a generated graph using that learned representation }
\end{figure*}

\newpage 

\section{Message Passing GVAE}

We introduce the Generative process and the decoder as well as the inference model and the encoder. 
Both the Encoder and Decoder use a message passing neural network (MPNN) to learn a graph structured representation during inference and generation. 

We use a variant of the MPNN from \cite{gilmer2017neural} that uses graph attention \cite{velivckovic2017graph} 
as an aggregation method and the message function similar to interaction networks \cite{battaglia2016interaction}.

The node update function uses a GRUCell \cite{chung2014empirical}.
After propagation through message passing layers, we use the set2set model \cite{vinyals2015order}
as the readout function to combine the node hidden features into a fixed-size graph level representation vector.

\subsection{Generative process}

In this section we describe the basic structure of the generative process starting with the underlying observation model and then discuss the structure of the decoder which consists of three main graph generation steps: 

1) Graph Read In 2) Message Passing and 3) Graph Readout. 

The basic structure of the generative mode, including these three steps, is visualized in Figure 2. 

\newpage 
\textbf{The observation model.} Based on the assumptions of the ER family, we assume the observation model factorizes:
\begin{align}
    p(G | \B z ) = \prod _{v\in G} p(\B x_v |\B z ) \prod _{u \in G } p(\B e _{uv} | \B z )
\end{align}
For molecular graphs we have that both edges and node features are categorical 
\begin{align}
  \B x_v \sim \texttt{Cat} (\B p_v) \text{ and } \B e _{uv} \sim \texttt{Cat} (\B p_{uv}) 
\end{align}

we also treat non-existent nodes and edges as a category. The three main steps the decoder uses to readout graphs from this distribution are 

\begin{enumerate}
    \item \textbf{Graph Read In.} The first step, reads in the initial graph representation from the fixed dimensional latent space by projecting the latent representation  with a linear layer then passing each vector through a RNN cell to construct an initial state for each node. Each edge is initialized with a zero vector. 
    \item \textbf{Message Passing.} Afterwards, using the initialized graph we perform message passing on both the node and edge representations. This allows us to create a representation we can use to read out a graph  
    \item \textbf{Graph Readout.} Lastly, using the final edge and node representation we transform the edge representation and predict independent edge and node probabilities. 
\end{enumerate}

\newpage 

\begin{algorithm}[tb]
\caption{Generating Graphs with MPGVAE}
\label{alg:algorithm}
\textbf{Input}: $G \sim p(\D_{\mathcal{G}})$ \\
\textbf{Initialize}: $\B \theta , \B \phi  $ 
\begin{algorithmic}[1] 
    \STATE $\{ \B h_v^T \}_{v\in G} = \texttt{MessagePassing}(\B x_v, \B e_{uv})$ 
    \STATE $\B \mu , \log \B \sigma  = \texttt{Set2Vec}(\{\B h^T_v, \B x_v\}_{v\in G} )$
    \STATE $\B z\sim q_{\B \phi}(\B z | G ) $ 
    \STATE $\{\B x^{(0)}_v, \B e^{(0)}_{uv}\} = \texttt{ReadIn}(\B z) $
    \STATE $\{\B x^{(T)}_v, \B e^{(T)}_{uv}\} = \texttt{MessagePassing}(\B x_v, \B e_{uv})$ 
    \STATE $\tilde G =  \texttt{ReadOut}(\{\B x^{(T)}_v, \B e^{(T)}_{uv}\}_{v\in G})$
\end{algorithmic}
\textbf{Return}  $ G \sim p_{\B \theta}(G |\B z) $ 
\end{algorithm}

\subsection{Inference Model.}
The variational posterior $ q(\B z  | G )$ is a factored Gaussian with mean vector $ \B \mu $ and
variance vector $\B \sigma ^2 $. We use a standard Gaussian prior $p(\B z) = \N(\B 0, \B I)$ .

\noindent \textbf{Encoder} The encoder is a MPNN  \cite{gilmer2017neural}
which takes in a molecular graph and encodes it into a fixed dimensional representation that is mapped to 
the parameters of the variational approximation. We describe the model below:

\noindent \textbf{Messages} We pass messages along edges where each message is updated using the 
current edge and node representation 
\begin{align}
    \B m^{t}_{uw} = \tanh (\B W _e \B e^{t}_{uw} + \B W _{h_u} \B h^{t}_{u}+ \B W _{h_w} \B h^{t}_{w})
\end{align}
this is a similar message function, to the one found in interaction networks \cite{battaglia2016interaction}.

\noindent \textbf{Edge Update} we use the message representation to directly update edges $  \B e^{t+1}_{uv} =  \B m^{t}_{uv} $ 
For all edges in the graph. The edges are initialized with the bond features.
\vspace{.1cm}

\noindent \textbf{Node Update} for each node we construct a message by aggregating using a graph attention \cite{velivckovic2017graph} where we first compute attention coefficients 
\begin{align}
    a _{uv} =  \exp(\B W \B m^t_{uv}) / \sum _{w\in \N_u} \exp(\B W \B m^t_{uw})
\end{align}
\noindent Then aggregate over neighbourhoods by summation : 
\begin{align}
    \B m^{(t+1)}_{uv} = \sum _{w\in \N_v}  a_{uv} \B m^t_{uw}  
\end{align}
Then each node is updated using the previous state and messages using a Gated Recurrent cell \cite{chung2014empirical} and \cite{li2015gated}
\begin{align}
    \B h _{v} ^{(t+1)} =  \texttt{\large GRUCell} (\B h _{v} ^{(t)} ,  \B m _{uv} ^{(t+1)}  )
\end{align}

\noindent \textbf{Read Out} After propagation through message passing layers, we use the set2set model \cite{vinyals2015order} as the readout function to combine the node hidden features into a fix-sized hidden graph level representation.
\begin{align}
 \B h_G =    \texttt{\large Set2Vec} (\{\B h^{(T)}_v\}_{v\in G})  
\end{align}

Using this representation, the mean and log variance of the variational posterior are computed with two separate linear layers. 
\begin{align}
    \B \mu = \ell_{\mu} \ (\B h_G) , \ \    \log \B \sigma  = \ell_{\sigma} (\B h_G) 
\end{align}

\newpage 

\subsection{Graph Read In and Message Passing}

\vspace{.15cm}

\textbf{Graph Read In}  First we read in the raw graph representation from the fixed dimensional latent space by projecting the latent representation to a high dimensional space with a linear layer with a sigmoid activation  $\sigma _z$. Then we pass each transformed vector through a RNN cell to read in an initial state for each node. Each edge is initialized with a zero vector. This is described in algorithm box 2. 

\vspace{.25cm}

\noindent \textbf{Message Passing} 
Using the initialized graph we perform message passing using both the node and edge representations.
We use a MPNN that is identical to the encoder's MPNN except does not perform any graph level aggregation. After $T$ propagation rounds we come to the final representation for each edge and node.

\begin{algorithm}[h]
\caption{\texttt{ReadIn} : From a latent sample we obtain an initial graph state to perform message passing on. }
\begin{algorithmic}[1]
\STATE \textbf{Input} $\B z  \sim q_{\phi} (\B z |G )$ 
\STATE $ \B h_z = \sigma ( \B z )$ 
\STATE $\{\B x^{(0)}_v \}_{v\in G} = \texttt{RNNCell}(\B h_ z) $
\STATE $ \{\B e^{(0)}_{uv} = \B 0 \}_{u,v\in G} $
\end{algorithmic}
\end{algorithm}

\subsection{Graph Readout}

Taking in the final edge and node representation we transform each edge representation 
such that we obtain a symmetric edge tensor-- simply by adding its transpose to itself and dividing by 2 thus making it symmetric. 

Then we use a linear layer given for nodes and edges  to map the learned representation 
from a high dimensional space back to the dimension space of the graph 
($\B e_{uv} $ has dimension \texttt{number of bond types} and each 
$\B x _v $ has dimension \texttt{num atoms}. 

From this state, using both representations we can sample independent edge and node probabilities using a softmax function to normalize the probabilities. 
\begin{align}
     \texttt{softmax}(\B x)_u = \exp(\B x_u)/ \sum_w \exp(\B x_w). 
\end{align}
\begin{algorithm}[h]
\caption{\texttt{ReadOut} graph $G = ( \B {\tilde  x}_v , \B {\tilde e} _{uv} )$}
\textbf{Input} $\{\B x^{(T)}_v, \B e^{(T)}_{uv}\} )$ \\
\textbf{For all} $u, v \in G$ \textbf{do} 
\begin{algorithmic}[1]
\STATE $\B {\tilde  x}_v    = \texttt{Softmax}(\sigma_x(\B x^{(T)}_v )  ) $
\STATE $\B e_{uv} = (\B e^{(T)}_{uv}+ \B e^{(T)}_{vu} ) /2 $
\STATE $\B {\tilde e} _{uv} = \texttt{Softmax}(\sigma_e(\B e_{uv} ) ) $
\end{algorithmic}
\textbf{Return} $ G = (\B {\tilde  x}_v , \B {\tilde e} _{uv}) $ 
\end{algorithm}

In the next section we discuss our experimental results.

\begin{figure*}
      \begin{center}
    \includegraphics[width=0.22\textwidth]{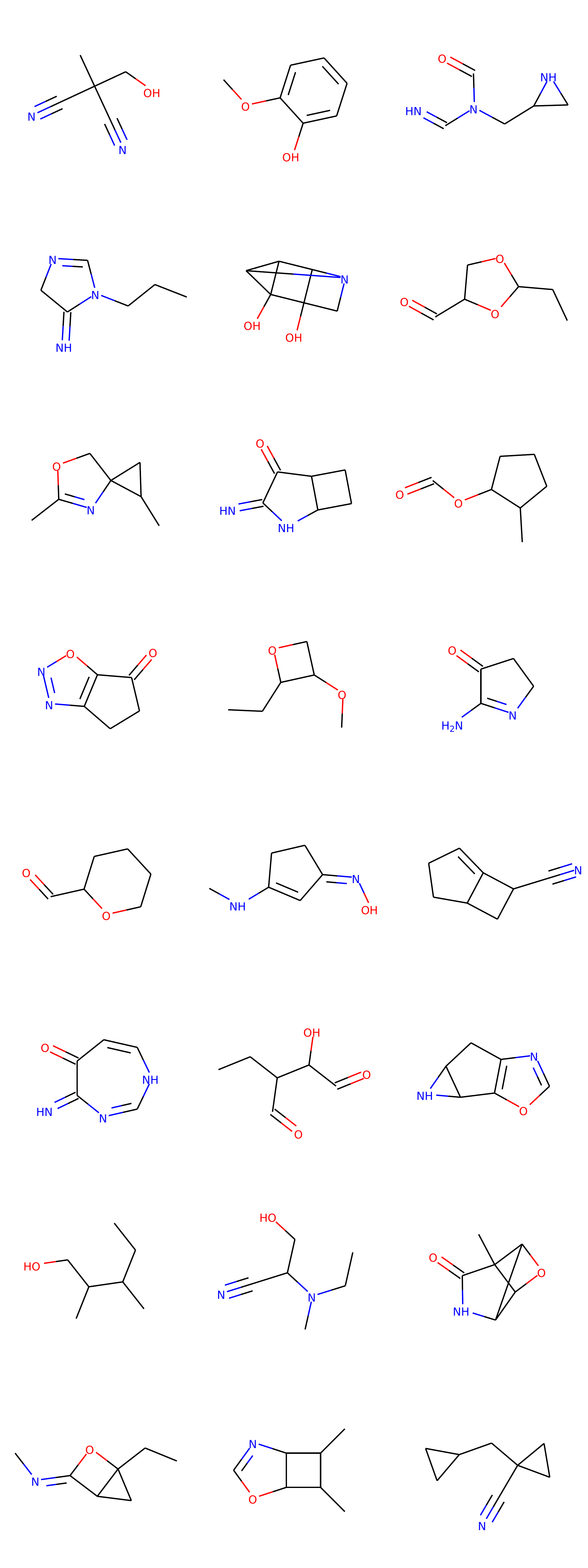}
    \hspace{.1cm}\vline\hspace{.1cm} 
    \includegraphics[width=0.22\textwidth]{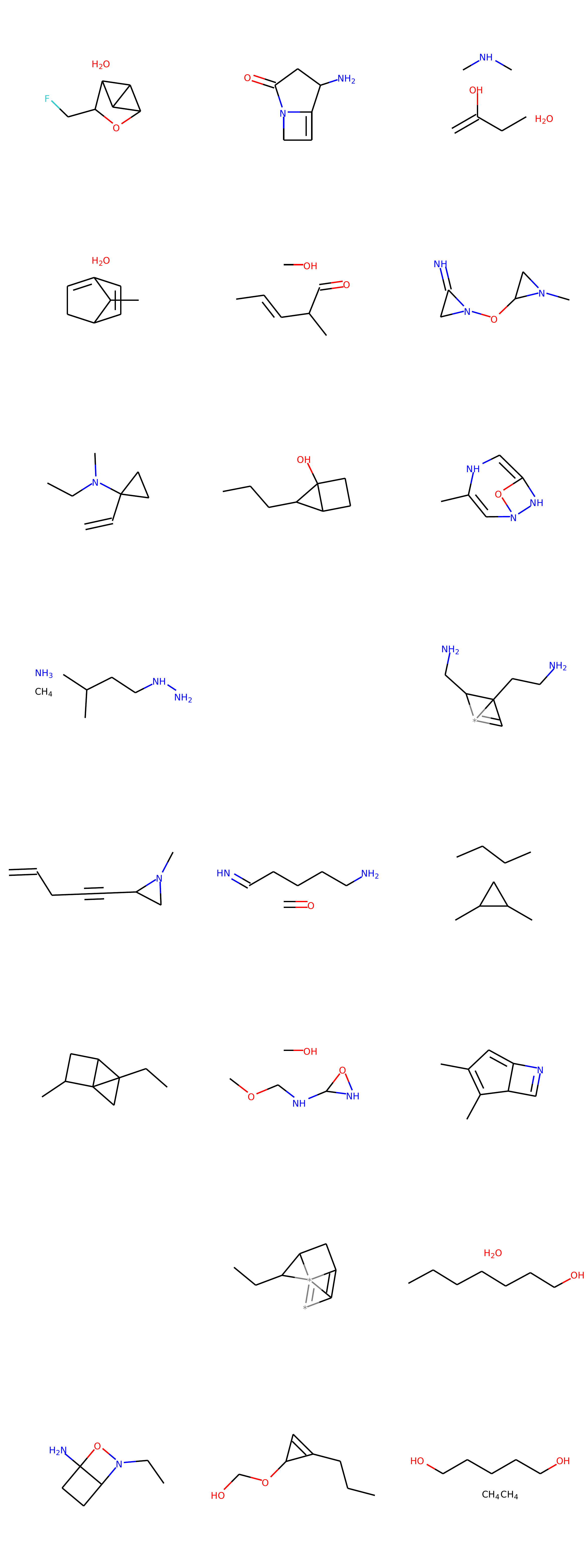}
    \hspace{.1cm}\vline\hspace{.1cm} 
    \includegraphics[width=0.22\textwidth]{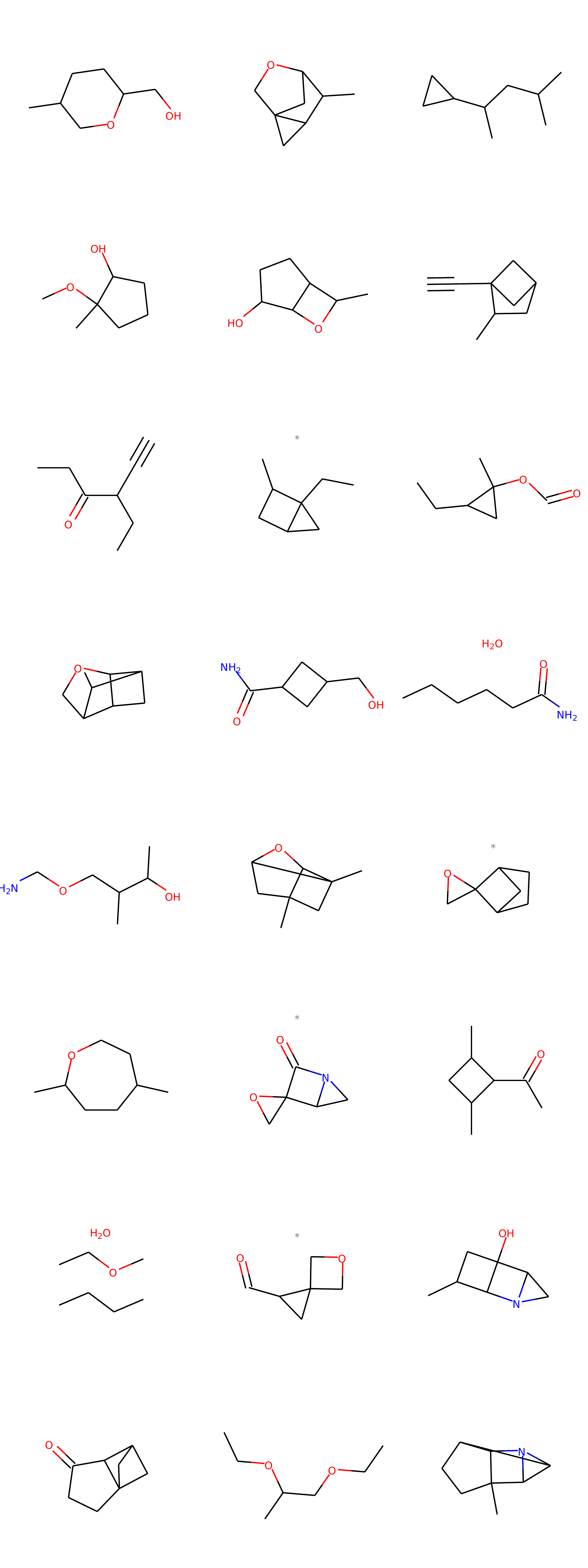}
    \hspace{.1cm}\vline\hspace{.1cm} 
    \includegraphics[width=0.22\textwidth]{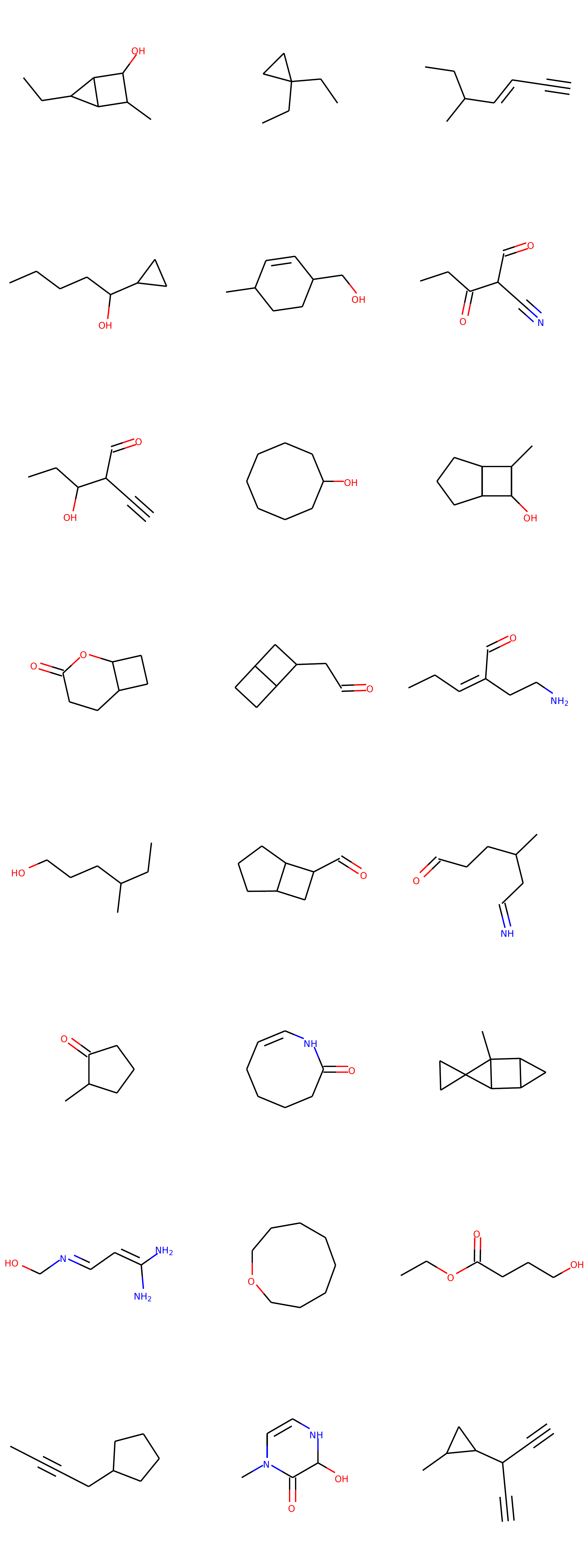}    
 \end{center}
\hspace{1cm}
\textbf{training samples} \hspace{2cm}
\textbf{gvae samples} \hspace{2cm}
\textbf{molgan samples} \hspace{2cm}
\textbf{mpgvae samples} \hspace{1cm}
\end{figure*}

\newpage

\section{Experiments}

We focus our evaluation efforts on a few standard tasks in molecular generative modeling following previous 
works \cite{gomez2018automatic} \cite{simonovsky2018graphvae} including :

\begin{itemize}
    \item \textbf{Molecular Generation Quality} We test the MPGVAE
     on the task of generating molecules when sampling from the prior distribution using a variety of
     established measures and plot samples from the model.
     \item \textbf{Matching The Training Data Distribution} We also test the MPGVAE on its 
     ability to match the distributional statistics of the training data.
     \item \textbf{Conditional Generation} Lastly, we test the ability of the MPGVAE to generate 
     molecules conditionally based on atom histograms.
\end{itemize}

\noindent \textbf{Dataset} In all experiments, we used QM9 \cite{ramakrishnan2015electronic} a subset of the massive 166.4 billion molecules
GDB-17 chemical database \cite{ruddigkeit2012enumeration}. QM9
contains 133,885 organic compounds of up to 9 heavy atoms:
carbon, oxygen, nitrogen and fluorine (CNOF).

\noindent \textbf{Baselines} We consider 2 main baselines GraphVAE \cite{simonovsky2018graphvae} and 
MolGAN \cite{de2018molgan}. We also compare with the Character VAE (CVAE)\cite{gomez2018automatic} and GrammarVae (GraVAE) \cite{kusner2017grammar} on the first task. 

\newpage

\begin{table}
\centering
\begin{tabular}{lrrrr}  
\toprule
Model  & Valid  & Unique & Novel & Num \\
\midrule
CVAE     & 0.10  & 0.68 & 0.90 & 612    \\
GraVAE     & 0.60  & 0.09 & 0.81 & 437    \\
GVAE & 0.81  & 0.24 & 0.61 & 1185    \\
MolGan   & 0.98  & 0.10 & 0.94 & 921    \\
\bottomrule
MPGVAE   & 0.91  & 0.68 & 0.54 & 3341    \\
\bottomrule
\end{tabular}
\caption{Comparison of Metrics}
\label{tab:booktabs}
\end{table}

\noindent \textbf{Model Configuration } Both the encoder and decoder MPNN use 4 layers with [32,64,64,128] units in the encoder and [64,64,32,32] in the decoder. The set2vec and vec2set functions use 256 hidden units in the graph level representation and latent space uses 18 dimensions. 

\subsection{Molecular Generation Quality} 

\noindent \textbf{Metrics} We generate $10^4$ samples from the prior 
and assess them using the following statistics defined in \cite{simonovsky2018graphvae}: 
1) Validity is the ratio between the number of valid and generated molecules. 
2) Uniqueness is the ratio between the number of unique samples and valid samples.
3) Novelty measures the ratio between the set of valid samples not in the training data and the total number of valid samples.

\begin{figure*}
\centering
    \includegraphics[width=0.32\textwidth]{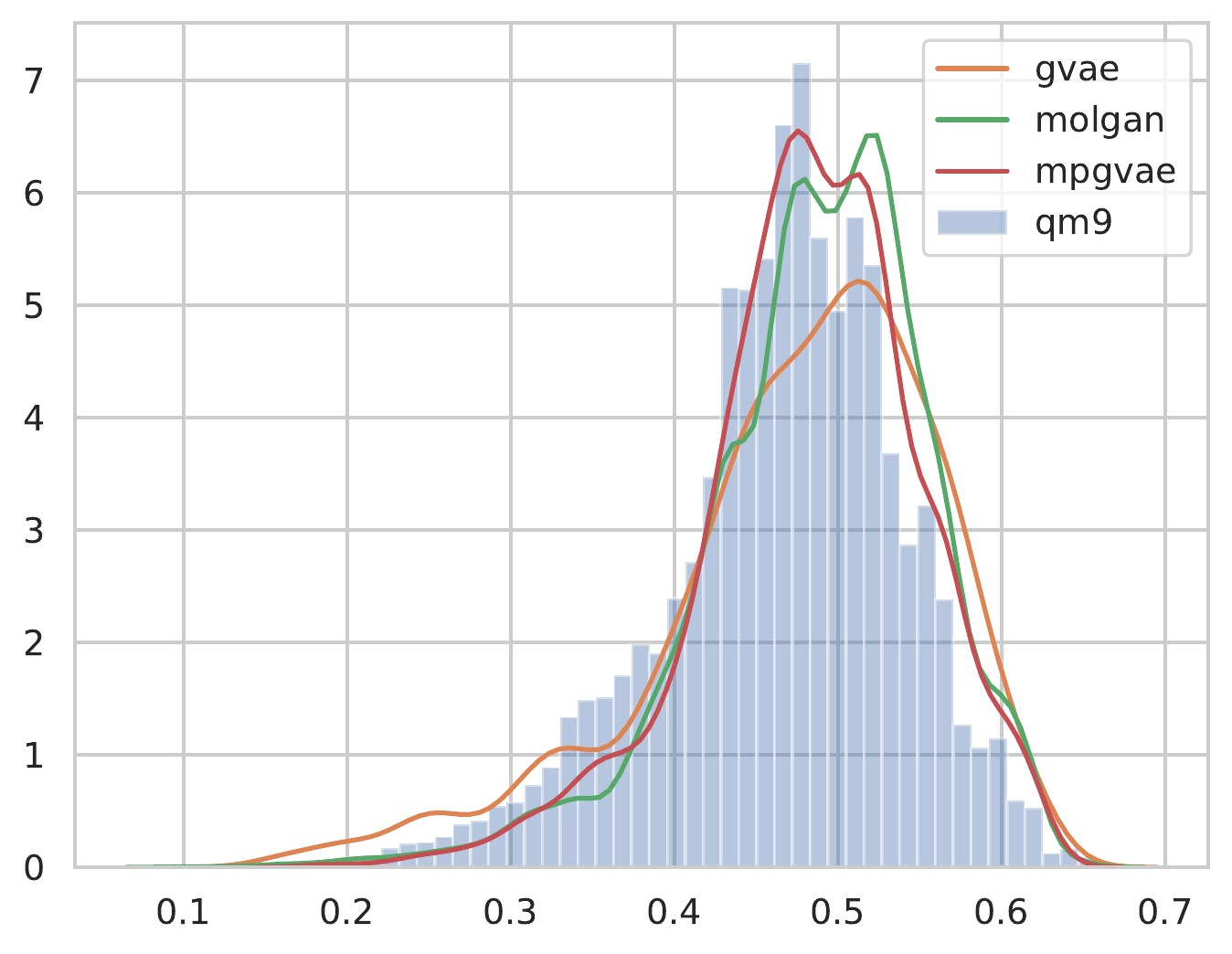}
    \includegraphics[width=0.32\textwidth]{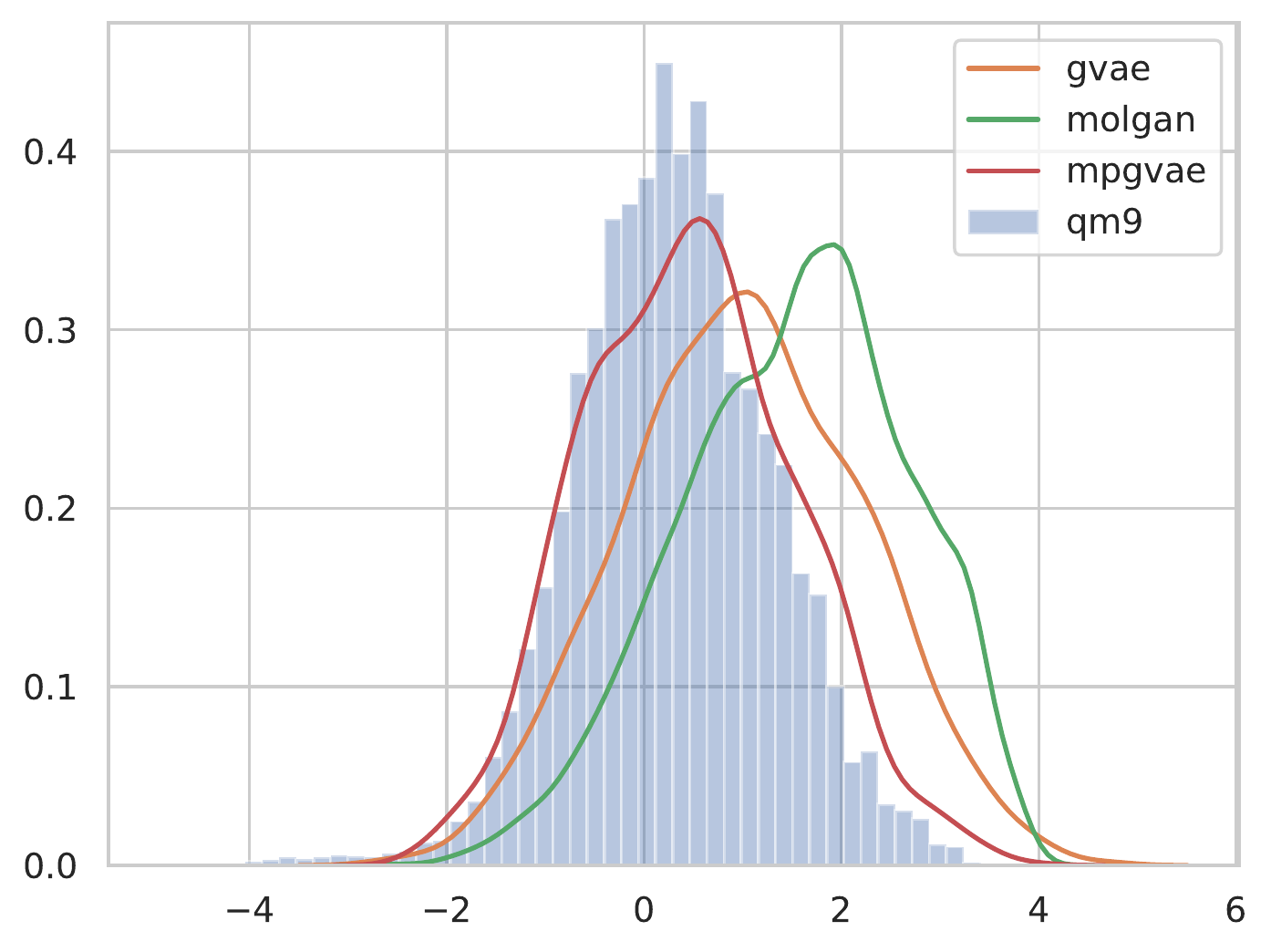}
    \includegraphics[width=0.32\textwidth]{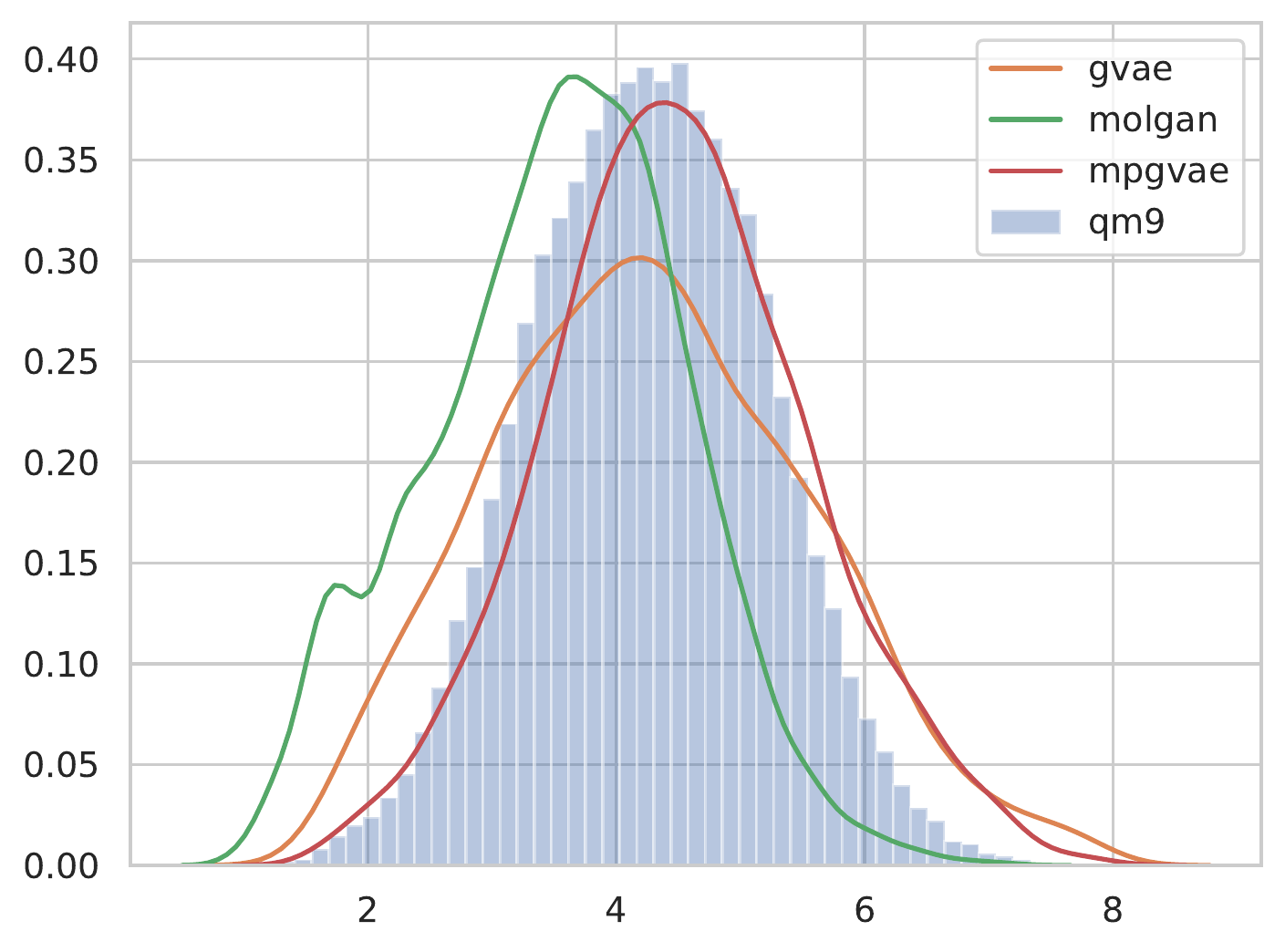}
    \caption{Distributions of QED (left), logp (middle), and SA (right) for sampled molecules and QM9}
\end{figure*}

\newpage 

We can see from table 1 that MPGVAE produces the best balance of the three measures while MolGAN, GVae and GraVAE cannot generate unique molecules --meaning they only learn to generate a few kinds of molecules -- leading to lower number of total valid, unique and novel generated (column Num). MPGVAE generates 300 percent more novel, valid unique molecules than GVAE- a substantial improvement. Also upon visual inspection MPGVAE generates more reasonable molecules than both GraVAE and MolGAN -- both of which generate a few molecules with disconnected, isolated scaffolding.

\subsection{Matching distributional statistics }

In this section, we use various statistics to compare our model’s learned
distribution to the training distribution 

\noindent \textbf{continuous statistics}.  Similar to \cite{seff2019discrete} we leverage three commonly used quantities when assessing  molecules: 
the quantitative estimate of drug-likeness (QED) score,  the synthetic accessibility (SA) score, and the log octanol-water partition coefficient (logP). Th metrics depend on many molecular features 
allowing for an overall comparison of distributional statistics.

\noindent \textbf{discrete statistics } we consider a second set of discrete statistics to measure how well each model captures the training distribution. Following \cite{liu2018constrained}, we measure the average number of each atom type in the generated samples, and count the average number of rings in each molecule. 

\noindent \textbf{Results} We evaluate the quality of our generative model by comparing the distribution that MPGVAE generates to those in the original data. In Fig. 3, we display Gaussian kernel density estimates (KDE) of the continuous metrics for generated sets of molecules from two baseline methods, in addition to MPGVAE. A normalized histogram of the QM9 training
distribution is also shown for visual comparison. In Fig. 4. we plot a stacked histogram of the average number of atoms and rings in each molecule. For each method, we use $10^4$ samples from the model. 

In figure 3, we see that MPGVAE more closely matches the distribution from QM9 for the continuous measures. For the discrete statistics, in figure 4 we see as well that MPGVAE's stacked barplot is closer to QM9's than the baselines. GVAE does not match the ring statistics while MPGVAE does. 

\begin{figure}[t]
    \centering
    \includegraphics[width=0.48\columnwidth]{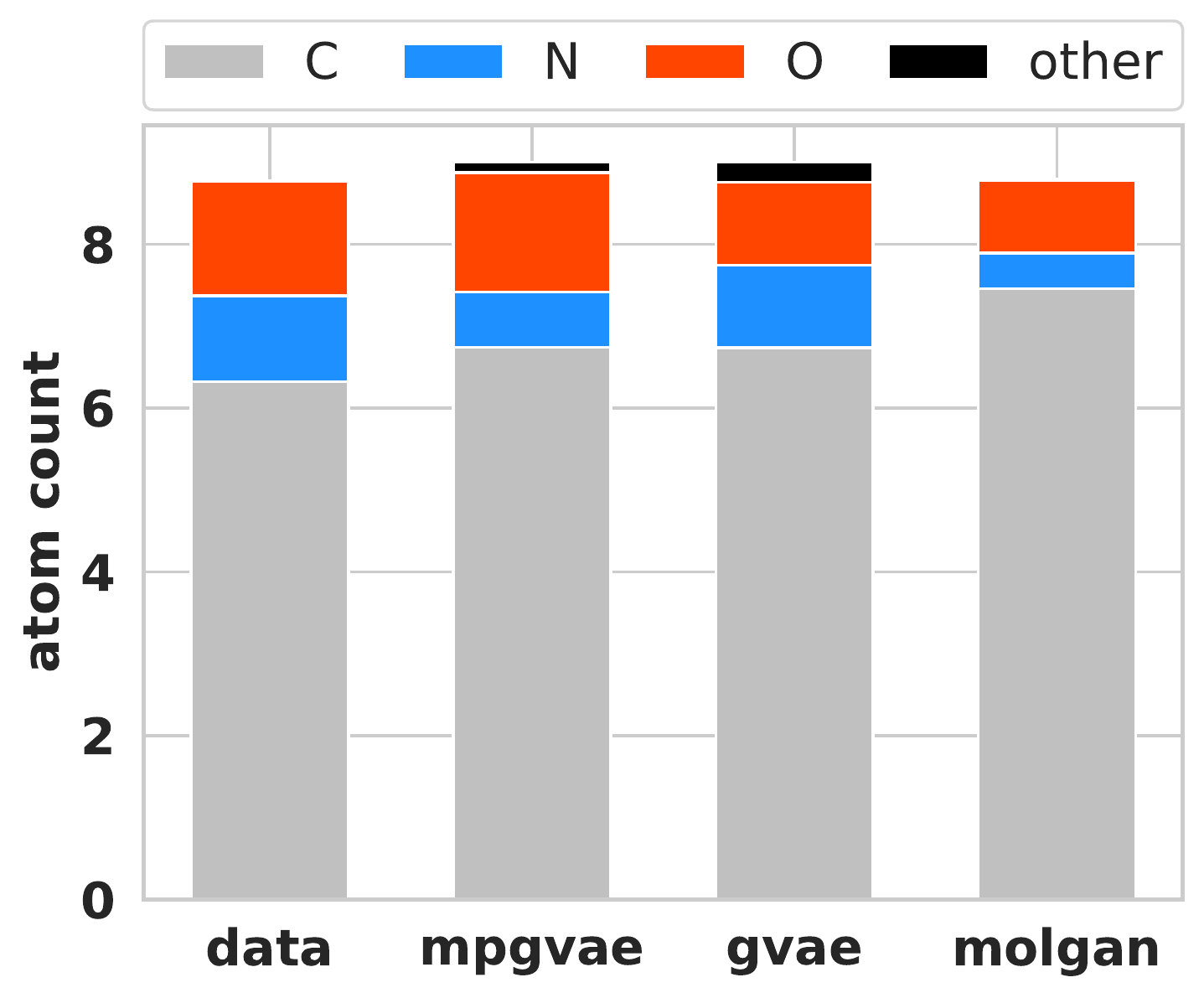}
    \includegraphics[width=0.48\columnwidth]{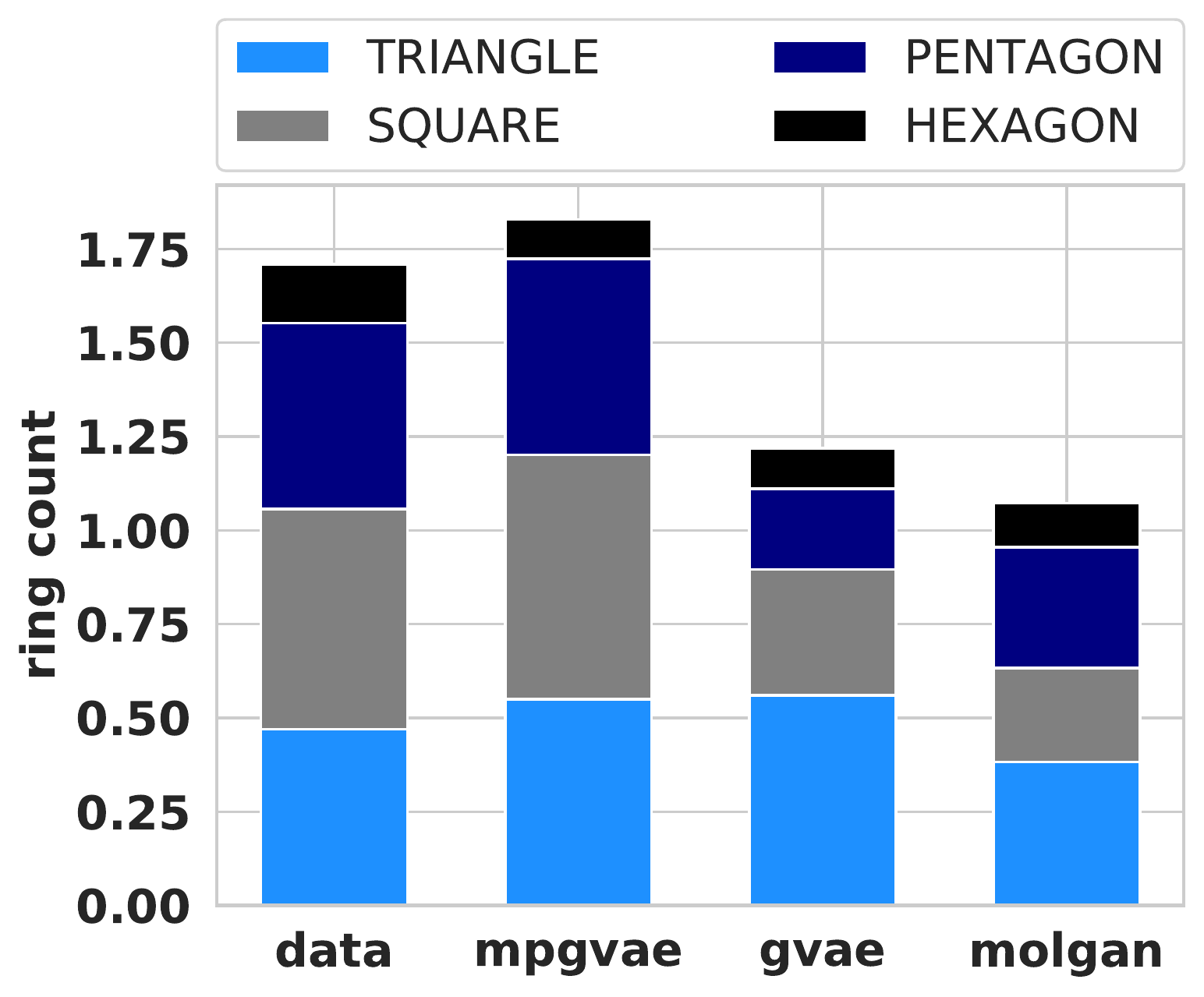}
    \caption{Distributions of atom, bond and ring count}
    \label{fig:my_label}
\end{figure}

\subsection{Conditional generation}

For more control over the generated molecules we can condition both encoder and decoder
on a label vector $\B y $ of atom histograms associated with each input molecule $G$ \cite{sohn2015learning}. The decoder $p(G|\B z,\B y)$ takes in $\B z$ and $\B y$, while in the encoder $\B y$ is concatenated to node features. 

Again we draw $10^4$ samples from the prior and as in  \cite{simonovsky2018graphvae}  compute the discrete point estimate of what is decoded $ \text{argmax} \  p (G|\B z , \B y ) $ 
We are interested in accuracy, which is the ratio of chemically valid molecules with atom histograms equal to their label $\B y$ over number of sampled molecules. We again see in table 2 that MPGVAE has improved accuracy over GVAE.

\begin{table}
\centering
\begin{tabular}{lrr}  
\toprule
Model  & Validity & Accuracy \\
\midrule
GVAE          & 0.57  & 0.47      \\
MPGVAE        & 0.89  & 0.67       \\

\bottomrule
\end{tabular}
\caption{Conditional Generation Task}
\label{tab:booktabs}
\end{table}

\section{Conclusion}

In this work we addressed the problem of generating graphs from within the ER family without edge correlation. 
We built a MPNN into both the decoder and encoder of a VAE and demonstrated it substantially improves 
GVAE. 

\newpage

\section*{Acknowledgments}

\appendix

\bibliographystyle{named}
\bibliography{ref}

\end{document}